\documentclass[review]{elsarticle}

\usepackage[utf8]{inputenc}	
\usepackage[T1]{fontenc}	
\usepackage{CJKutf8}	

\usepackage{lineno,hyperref}
\modulolinenumbers[5]

\usepackage{adjustbox}	
\usepackage{booktabs}	
\usepackage{float}	
\usepackage{makecell}	
\usepackage{multirow}   
\usepackage{natbib}	
\usepackage{orcidlink}	
\usepackage{subcaption}	
\usepackage{tabularx}	
\usepackage{tabulary} 
\usepackage{utfsym}	

\usepackage{geometry}
\usepackage{setspace}
\usepackage[flushleft]{threeparttable}

\usepackage{fancyhdr}
\usepackage[section]{placeins}	

\biboptions{authoryear}

\begin{document}
\begin{CJK}{UTF8}{gbsn}	

\begin{frontmatter}


\title{Large language models for automated scholarly paper review: A survey}



\author[mymainaddress]{Zhenzhen Zhuang}
\ead{zhuangzz@gzist.edu.cn}

\author[mymainaddress]{Jiandong Chen}
\ead{jiandongchen@gzist.edu.cn}

\author[mysecondaryaddress]{Hongfeng Xu}
\ead{hfxu1977@gznu.edu.cn}

\author[mythirdaddress]{Yuwen Jiang}
\ead{jiangyuwen@gzist.edu.cn}

\author[mymainaddress]{Jialiang Lin\corref{mycorrespondingauthor}}
\cortext[mycorrespondingauthor]{Corresponding author}
\ead{me@linjialiang.net}

\address[mymainaddress]{School of Computer Science and Engineering, Guangzhou Institute of Science and Technology, Guangzhou, China}
\address[mysecondaryaddress]{School of Economics and Management, Guizhou Normal University, Guiyang, China}
\address[mythirdaddress]{School of Artificial Intelligence, Guangzhou Institute of Science and Technology, Guangzhou, China}

\begin{abstract}

Large language models (LLMs) have significantly impacted human society, influencing various domains. Among them, academia is not simply a domain affected by LLMs, but it is also the pivotal force in the development of LLMs. In academic publication, this phenomenon is represented during the incorporation of LLMs into the peer review mechanism for reviewing manuscripts. LLMs hold transformative potential for the full-scale implementation of automated scholarly paper review (ASPR), but they also pose new issues and challenges that need to be addressed. In this survey paper, we aim to provide a holistic view of ASPR in the era of LLMs. We begin with a survey to find out which LLMs are used to conduct ASPR. Then, we review what ASPR-related technological bottlenecks have been solved with the incorporation of LLM technology. After that, we move on to explore new methods, new datasets, new source code, and new online systems that come with LLMs for ASPR. Furthermore, we summarize the performance and issues of LLMs in ASPR, and investigate the attitudes and reactions of publishers and academia to ASPR. Lastly, we discuss the challenges and future directions associated with the development of LLMs for ASPR. This survey serves as an inspirational reference for the researchers and can promote the progress of ASPR for its actual implementation.

\end{abstract}


\end{frontmatter}


\section{Introduction}
\label{sec:intro}

In November 2022, OpenAI introduced ChatGPT, a groundbreaking AI-powered chatbot.\footnote{https://openai.com/index/chatgpt/} This chatbot demonstrated an exceptional ability to respond to a wide range of queries with high-quality outputs. Its capabilities include integrating and summarizing literature, assisting with student compositions, composing musical scores, generating scripts for poems and lyrics, writing and debugging code, emulating various terminals, and even simulating Linux systems. ChatGPT gained immense popularity, reaching 100 million users within just two months of its release~\citep{wu-brief-2023}, making it the fastest-growing consumer application in history at the time. Bill Gates remarked that ChatGPT's impact is comparable to the invention of the internet, emphasizing that it ``will change our world''.\footnote{https://www.reuters.com/technology/microsoft-co-founder-bill-gates-chatgpt-will-change-our-world-2023-02-10/} At the heart of ChatGPT lies large language models (LLMs), an advanced artificial intelligence (AI) technology that has rapidly gained public attention. Within just two years, LLMs have begun transforming human lifestyles and professional workflows. This breakthrough has ushered in significant advancements in the fields of natural language processing (NLP) and AI, marking a new era of innovation and reshaping the way society interacts with technology.

LLMs~\citep{zhao-survey-2023} are advanced deep neural network language models characterized by an extensive number of parameters, which are optimized through training on large amounts of text data. These models excel at capturing and modeling the complexity and diversity of natural language, enabling them to process and generate text with a high degree of fluency and coherence that closely resembles human language use~\citep{brown-language-2020,kasneci-chatgpt-2023,distefano-exploring-2024}. The term ``large'' in LLMs refers to their immense scale, with model sizes ranging from hundreds of millions to hundreds of billions of parameters. These models are trained on vast amounts of textual data, allowing them to learn intricate linguistic patterns and structures. As the scale of an LLM increases, so does its ability to comprehend nuanced language and perform sophisticated language-related tasks. With advancing technology, LLMs now support multimodal capabilities, handling text, audio, images, and more. OpenAI's GPT-4o~\citep{openai-gpt4o-2024}, for instance, demonstrates advanced real-time reasoning and sensory understanding across modalities. These enhanced abilities improve user interaction and broaden applicability across industries.

In the domain of peer review, research indicates that, LLMs have already been employed to generate the content of review report~\citep{liang-monitoring-2024}. As LLMs continue to evolve, their influence on this domain is inevitable, necessitating a thorough examination of their role in scholarly evaluation. In our previous work, we proposed the concept and pipeline of automated scholarly paper review (ASPR) and outlined its potential future development~\citep{lin-automated-2023}. At that time, ChatGPT had just been released, and we explicitly stated that the future directions for ASPR should ``embrace the changes brought by ChatGPT and other large language models (LLMs)''. Since then, LLMs have been increasingly utilized by researchers to conduct ASPR or perform similar tasks. Furthermore, various studies, though not originally designed for ASPR, have demonstrated applicability in this domain, reinforcing our previously proposed concept of the coexistence phase where ASPR and traditional peer review function alongside each other. At this stage, ASPR serves as a human assistant to enhance the quality and efficiency of peer review. As its adoption grows, advancements in ASPR-related technologies, resource availability, policies, and ethical considerations continue to evolve, drawing increasing attention from the research community. Given the significance of these developments to scholars worldwide, it is crucial to systematically collect, analyze, and review this emerging body of work.

Accordingly, we undertake this survey to provide a comprehensive review of the state of ASPR in the era of LLMs. The research question of this survey centers on how LLMs promote the advancement of ASPR and related topics.
A snowballing method was employed to select the relevant literature, with five papers (\citet{hosseini-fighting-2023}, \citet{robertson-gpt4-2023}, \citet{biswas-chatgpt-2023}, \citet{liu-reviewergpt-2024}, \citet{liang-can-2024}) selected as the seeds. While some of these papers were initially identified using their arXiv versions, the formally published versions are cited in this paper. Following the procedure outlined by \citet{wohlin-guidelines-2014}, the collection was expanded through both forward and backward citation searches, only including papers that met our defined criteria. First, the paper must have been made publicly available between 2023 and 2024, including submissions to arXiv. This timeframe corresponds to the era of LLMs, which is recognized to have matured with the release of ChatGPT. To account for possible delays between research completion and publication, the starting point was set at 2023 and the endpoint at 2024, aligning with the period during which this survey was conducted. Second, to meet the research question, we only included papers that utilized LLMs for conducting ASPR. Additionally, supporting literature was provided for technical terminology, historical context and publisher polices, etc.

Also, we would like to note that issues such as potential biases, fairness, ethics, accountability, and the risk of misuse have been discussed in our previous work~\citep{lin-automated-2023}. While the emergence of LLM-driven ASPR represents a technological advancement, the core considerations surrounding these issues remain largely unchanged. In light of this, to maintain focus and avoid redundancy, we refer readers to our earlier study for an in-depth discussion of these issues.

This survey is structured as follows. In Section~\ref{sec:llm-usage}, we start by reviewing which LLMs researchers have used to conduct ASPR. In Section~\ref{sec:new-tech}, we analyze how LLMs address the technological bottlenecks for ASPR. In Section~\ref{sec:new-method}, we review what new methods are employed in LLMs for ASPR. In Section~\ref{sec:new-dataset} and \ref{sec:new-code-system}, we summarize what new LLM resources are available for ASPR. In Section~\ref{sec:llm-performance} and \ref{sec:existing-issue}, we examine how LLMs are performing in ASPR, highlighting both strengths and existing issues. In Section~\ref{sec:publisher-policy} and \ref{sec:academia-suggestion}, we assess how publishers and academia perceive and react to LLMs for ASPR. In Section~\ref{sec:concl}, we conclude this survey by outlining what the challenges and where the future directions are for LLMs in ASPR.

\section{Background of LLMs and their applications in ASPR}
\label{sec:llm-usage}

\subsection{Development process of large language models}

The technological development of LLMs has followed an evolutionary trajectory progressing from shallow statistical modeling~\citep{ai-enhancement-2024} to deep semantic understanding~\citep{shen-efficient-2024,johannes-survey-2024}, marked by four distinct paradigm shifts. Early efforts focused on statistical language models based on the Markov assumption such as n-gram models, achieving basic language prediction through local word order probability modeling~\citep{anderson-more-1972}. This was followed by the emergence of neural language models which introduced recurrent neural networks (RNNs) and their variants, such as long short-term memory (LSTM) networks. These architectures enabled the first effective distributed representation learning of text sequences~\citep{bengio-nerual-2003}. The advent of pre-trained language models, built on the Transformer architecture~\citep{vaswani-attention-2017}, exemplified by BERT~\citep{devlin-bert-2019} and the GPT series~\citep{radford-improving-2018}, further advanced the field by overcoming limitations in context awareness through self-supervised pre-training. Most recently, modern LLMs have pushed the frontier by scaling parameter counts to hundreds of billions, giving rise to emergent capabilities driven by quantitative-to-qualitative transformations and scaling laws~\citep{brown-language-2020}. Research indicates that once critical thresholds in model size, typically ranging from $10^9$ to $10^{12}$ parameters, and training data volume are surpassed, LLMs begin to demonstrate complex capabilities, such as in-context learning, instruction following, and chain-of-thought reasoning~\citep{kaplan-scaling-2020,liang-holistic-2023}. For instance, GPT-3, with 175 billion parameters, has achieved cross-task generalization in few-shot learning scenarios through meta-learning mechanisms. Similarly, Llama 3, by incorporating 17\% structured code data into its pre-training corpus, has improved its zero-shot logical reasoning capability by 23.6\% compared to its predecessor~\citep{touvron-llama-2023,liang-holistic-2023}. Nevertheless, with the continued in-depth study of LLMs, it has become evident that models such as GPT-3 and GPT-4~\citep{openai-gpt4-2023} still exhibit limitations in deep reasoning capabilities. Consequently, the focus of AI research and development has progressively shifted from pattern recognition to reasoning ability enhancement, with the objective of achieving a more logical and human-like thinking process. In December 2024, OpenAI launched the o1 reasoning models~\citep{openai-o1-2024,wu-comparative-2024}, and shortly thereafter in January 2025, DeepSeek introduced DeepSeek-R1~\citep{deepseek-r1-2025}, marking a significant step forward in this direction. These next-generation LLMs have demonstrated significant breakthroughs in addressing complex problems, thereby narrowing the gap between LLMs and human cognition. Among these advancements, DeepSeek-R1 stands out due to its exceptional cost-effectiveness and open-source design. By challenging traditional paradigms of AI development and facilitating broader access to cutting-edge models, DeepSeek-R1 represents a pivotal advancement for AI in the simulation of human cognitive patterns.

However, many application scenarios, such as edge devices, mobile platforms, and offline environments, are constrained by limited resources. To address the high computational demands and deployment challenges of LLMs, researchers have employed model compression techniques to reduce model size while maintaining functionality. These compressed models utilize only a small fraction of the parameters present in LLMs, which typically contain millions to billions of parameters~\citep{matarazzo-survey-2025}. Several strategies have been employed in this lightweight design to enable efficient operation in resource-constrained settings. Pruning and quantization are commonly used to minimize model redundancy, while knowledge distillation enables effective knowledge transfer within a teacher-student framework. In addition, parameter-efficient dynamic inference architectures have been developed to significantly reduce computational demands and improve inference speed. Experimental results validate the effectiveness of these strategies. For example, DistilBERT retains 97.2\% of the BERT-base's performance on the GLUE benchmark while achieving a 40\% reduction in parameter count through hierarchical distillation~\citep{sanh-distilbert-2019}. Furthermore, lightweight models, such as Gemma-7B, have demonstrated superior performance on tasks like GSM8K, a benchmark for mathematical reasoning, compared to other models of similar size, such as Llama 2-7B and Mistral-7B. This performance advantage is largely attributed to optimized attention mechanisms~\citep{gemma-gemma-2024}.

\subsection{Core components of large language models}

The core components of LLMs encompass fundamental techniques, architectural design principles, training strategies, and adaptation methods. In this subsection, we provide a high-level overview of these components, and refer readers to the original research literature for more detailed discussions.

The fundamental techniques include tokenization, position encoding, attention mechanisms, activation functions, and normalization strategies. Tokenization methods, such as Byte Pair Encoding (BPE)~\citep{sennrich-neural-2016} and WordPiece~\citep{schuster-japanese-2012}, convert text into discrete tokens, forming a foundation for subsequent processing. Positional encoding conveys sequence order information to the Transformer using techniques such as Attention with Linear Biases (ALiBi)~\citep{press-train-2022} and Rotary Position Embedding (RoPE)~\citep{su-roformer-2024}. These methods enhance the model's ability to process long sequences, with RoPE, in particular, offering a novel approach by encoding relative position through vector rotations. Attention mechanisms, such as sparse attention~\citep{child-generating-2019}, which utilizes sparse factorizations of the attention matrix to restrict computations to local contexts, and fast attention~\citep{dao-flashattention-2022}, which leverages memory optimization techniques to improve efficiency, have been proposed to address the quadratic computational complexity of standard self-attention over long sequences. Activation functions, such as GeLU~\citep{hendrycks-gaussian-2018} and variants of GLU~\citep{shazeer-glu-2020}, further enhance the model's capacity for nonlinear representation. Finally, normalization strategies, including pre-layer normalization~\citep{baevski-adaptive-2019} and DeepNorm~\citep{wang-deepnet-2024}, optimize network architecture by stabilizing training.

Currently, mainstream LLMs are predominantly based on the Transformer, with their core designs systematically classified into three architectures. The Encoder-Decoder architecture is designed primarily for sequence-to-sequence tasks. This architecture employs a distinct encoder to process the input sequence and a decoder to perform autoregressive output generation, enabling bidirectional context modeling and conditional generation, with T5~\citep{raffel-exploring-2020} serving as a representative example. The Decoder-Only architecture utilizes a masked autoregressive mechanism that predicts tokens based on preceding context to achieve open-domain text generation. This architecture has become the dominant choice for modern LLMs due to its computational efficiency, exemplified by the GPT series~\citep{radford-improving-2018}. The Encoder-Only architecture relies on bidirectional attention to capture global semantic relationships. This architecture is primarily suited for tasks such as text classification that do not require text generation, with BERT~\citep{devlin-bert-2019} being a prominent example.

In recent years, the Mixture-of-Experts (MoE)~\citep{shazeer-outrageously-2017} has been introduced to overcome the model scale bottleneck, serving as an expansion architecture. By dynamically activating sparse expert sub-networks, MoE enhances model capacity while maintaining computational efficiency. For instance, Gemini 1.5~\citep{xue-wdmoe-2024} combines MoE with the Decoder-Only architecture to form MoE-Decoder, which is capable of handling complex inference tasks by activating only about 10\% of its parameters. In contrast, mainstream models, such as GPT-4~\citep{openai-gpt4-2023}, Llama 3~\citep{touvron-llama-2023}, Qwen2~\citep{yang-qwen2-2024}, and Mistral 7B~\citep{jiang-mistral-2023}, continue to rely on the pure Decoder-Only architecture, balancing performance and generalization through autoregressive generation. The architectural details of other LLMs are summarized in Table~\ref{tab:llms-collection-aspr}. It is evident that the overall technical ecosystem remains dominated by the Decoder-Only architecture.

In terms of training strategies, distributed training facilitates ultra-large-scale training through the integration of data, tensor, and model parallelism~\citep{rajbhandari-zero-2020}. Libraries, such as Megatron-LM~\citep{shoeybi-megatron-2019} and DeepSpeed~\citep{rasley-deepspeed-2020}, have demonstrated significant advancements in parameter synchronization and memory optimization. Complementing these strategies, data preprocessing plays a crucial role in ensuring training quality, incorporating techniques such as quality filtering to remove low-quality text~\citep{mielke-between-2021}, deduplication to mitigate memory bias~\citep{chen-mxnet-2015}, and privacy filtering to prevent the leakage of sensitive information. Finally, the pre-training objectives, typically involving autoregressive modeling, masked language modeling~\citep{wang-what-2022}, and unified language modeling~\citep{dong-unified-2019}, are designed to equip models with universal semantic representations.

During the adaptation phase, fine-tuning techniques are employed to enhance practical applicability of LLMs. One common approach is instruction fine-tuning methods such as Flan~\citep{chung-scaling-2024}, which leverages multi-task data to improve zero-shot generalization. Another key approach is alignment fine-tuning, which utilizes reinforcement learning from human feedback to ensure that the model adheres to the standards of being ``useful, honest, and harmless''~\citep{ziegler-fine-2019}. It works by generating unexpected responses, then updating the model's parameters to avoid such occurrences. Additionally, parameter-efficient fine-tuning methods, including LoRA~\citep{hu-lora-2022} and adapters~\citep{houlsby-parameter-2019}, have been proposed to reduce the costs associated with fine-tuning by updating only a small fraction (typically between 0.01\% to 3\%) of the model's parameters.

\subsection{Application and comparative analysis of LLMs in ASPR}

In this subsection, we conduct a comprehensive review of the literature on the application of LLMs in ASPR. These LLMs can be categorized into two main groups: open-source and closed-source LLMs. These two categories differ in several key areas, including transparency, community ecology, customization, flexibility, and performance. Given the large number of LLMs available for ASPR, this paper mainly presents a comparative analysis of three representative open-source LLMs: Llama 3~\citep{touvron-llama-2023}, Mistral~\citep{jiang-mistral-2023} and Qwen2~\citep{yang-qwen2-2024}, as well as three prominent closed-source LLMs: GPT-4~\citep{openai-gpt4-2023}, Gemini 1.5~\citep{gemini-gemini-2024} and Claude 3~\citep{anthropic-claude-2024}.

Open-source LLMs demonstrate significant advantages in terms of transparency and accessibility, providing a more open platform that fosters academic research and technological innovation. The modular design of these models allows researchers to conduct in-depth analyses of their internal mechanisms. This openness promotes collaborative innovation among the global developer community. In contrast, closed-source LLMs operate largely as ``black boxes''. The opacity surrounding critical information, such as training data selection criteria and model fine-tuning strategies, severely limits the feasibility of external audits and ethical reviews~\citep{manchanda-open-2024}. This information closure not only hinders cross-institutional collaboration and technological democratization, but also raises significant concerns regarding the reliability of model outputs, algorithmic fairness, and potential social impacts. From research ethics to technological application, the closed-source strategy is becoming a key bottleneck that restricts the sustainable development of LLMs.

Customization and flexibility are the core advantages of open-source LLMs. Open-source LLMs offer users the ability to deeply customize and optimize models to meet specific needs, seamlessly integrate them into diverse systems, and substantially reduce both development and deployment costs. In contrast, closed-source LLMs primarily provide standardized API functionalities, with limited opportunities for customization. This lack of flexibility not only increases usage costs but also restricts innovation tailored to specific use cases. The open-source ecosystem promotes the rapid proliferation of new technologies through continuous iteration and sharing mechanisms, creating a virtuous cycle of demand, innovation, and application. Conversely, the innovation pace in the closed-source ecosystem is constrained by the update schedules set by the suppliers.

In terms of performance and efficiency, closed-source LLMs currently maintain a leading position. According to the authoritative evaluation list, Chatbot Arena~\citep{chiang-chatbot-2024}, closed-source LLMs have achieved superior results across multiple benchmark tests, including MMLU~\citep{wang-mmlu-2024} and HumanEval~\citep{zhang-humaneval-2025}. However, open-source LLMs are rapidly catching up. For example, Llama 3 is approaching the performance level of GPT-4 in code generation tasks, and its 70B model has become an affordable alternative to GPT-3.5 for enterprises. In the context of reviewing, experiments indicate that when LLMs act as meta-reviewers, closed-source LLMs (GPT-4, Claude 3, Gemini 1.5) are generally superior to their open-source counterparts (Llama 3, Qwen2) in identifying defective segments. Despite this, the F1 scores for all models remain relatively low, with recall scores surpassing precision scores, indicating a tendency toward false positives~\citep{du-llms-2024}.

Table~\ref{tab:llms-collection-aspr} presents an overview of various LLMs employed in ASPR applications, detailing their model names, open-source status, architecture types, official websites, and citing papers. The open-source status of LLMs is shown in the ``OS'' column, with a tick indicating open-source models and a cross indicating closed-source models. The architecture types are categorized as Encoder-Decoder, Decoder-Only, Encoder-Only, and MoE-Decoder, which are abbreviated as Enc-Dec, Dec-Only, Enc-Only, and MoE-Dec, respectively.

To further deepen the understanding, Figure~\ref{fig:llms-collection-aspr} visualizes the content in Table~\ref{tab:llms-collection-aspr}, presenting an overview in a clearer graphical manner. In Figure~\ref{fig:llms-collection-aspr}, the horizontal bar chart illustrates the frequency with which various LLMs appear in the ASPR literature. The chart encompasses both closed-source and open-source LLMs, as detailed in Table~\ref{tab:llms-collection-aspr}. The length of each bar signifies the number of times the corresponding model is cited in the literature. Notably, GPT-4 is the most frequently referenced, constituting nearly one-third of the proportion, followed by ChatGPT at about one-tenth of the share. In contrast, each of the remaining models accounts for only a single-digit percentage. This distribution reveals a significant disparity between the use of GPT-4 and ChatGPT versus other models, with the former two dominating the literature in terms of frequency. However, beyond the most widely used models, numerous alternative options are available, allowing users to select the most suitable tool for their specific needs. This demonstrates that LLMs exhibit significant diversity in the field of ASPR.

The pie chart of ``Source Type'' demonstrates the distribution between open-source and closed-source models. Open-source models account for approximately 2.7 times the percentage of closed-source models. This dominance of open-source solutions underscores the collaborative spirit and innovative drive among researchers and developers in advancing ASPR-related methodologies. Open-source models are particularly valued for their transparency and verifiability. Their internal mechanisms are publicly accessible, which facilitates a comprehensive understanding and independent verification of the research results. This openness ensures transparency, reproducibility, and adherence to the stringent requirements of peer review. In addition, the absence of license fees makes open-source models ideal for resource-limited settings.

\begin{table}[H]
	\centering
	\caption{Summary of LLMs used in ASPR applications}
	\label{tab:llms-collection-aspr}
	
	\setcellgapes{1pt}
	\makegapedcells
	\scriptsize
	\setlength{\tabcolsep}{2pt} 
	\renewcommand{\arraystretch}{1.3}
	\begin{adjustbox}{center, width=0.96\textwidth} 
		\begin{tabular}{p{80pt}lllp{80pt}}
			\hline
			Name  & OS  & Architecture & Official website & Citing papers \\ \hline 
			\makecell[l]{Alpaca \\ \citep{taori-alpaca-2023}}
			& \makecell[l]{\usym{2713}} & Dec-Only & \makecell[l]{\href{https://github.com/tatsu-lab/stanford_alpaca}{\texttt{github.com/tatsu-lab/stanford\_alpaca}}} & \makecell[l]{\citet{liu-reviewergpt-2024}}  \\
			\hline
			\makecell[l]{Bard \\ \citep{pichai-important-2023}}
			& \makecell[l]{\usym{2717}} & Dec-Only & \makecell[l]{\href{https://bard.google.com}{bard.google.com}} & \makecell[l]{\citet{liu-reviewergpt-2024}}  \\
			\hline
			\makecell[l]{ChatGPT  \\ \citep{openai-introducing-2022}}
			& \makecell[l]{\usym{2713}} & Dec-Only & \makecell[l]{\href{https://openai.com/index/chatgpt/}{openai.com/index/chatgpt/}} &\makecell[l]{\citet{biswas-ai-2024},\\ \citet{chawla-is-2024},\\ \citet{mehta-application-2024},\\ \citet{saad-exploring-2024},\\  \citet{zhou-is-2024}} \\
			\hline
			\makecell[l]{Claude 3  \\ \citep{anthropic-claude-2024}}
			& \makecell[l]{\usym{2717}} & MoE-Dec & \makecell[l]{\href{https://www.anthropic.com/news/claude-3-family}{www.anthropic.com/news/claude-3-family}} & \makecell[l]{\citet{du-llms-2024},\\  \citet{idahl-openreviewer-2024},\\ \citet{taechoyotin-mamorx-2024},\\  \citet{tyser-ai-2024}}   \\
			\hline
			\makecell[l]{Command R+ \\ \citep{gomez-introducing-2024}}
			& \makecell[l]{\usym{2713}} & Dec-Only & \makecell[l]{\href{https://huggingface.co/CohereForAI/c4ai-command-r-plus}{huggingface.co/CohereForAI/c4ai-command-r-plus}} & \makecell[l]{\citet{tyser-ai-2024}} \\
			\hline
			\makecell[l]{Dolly \\ \citep{conover-hello-2023}}
			& \makecell[l]{\usym{2713}} & Dec-Only & \makecell[l]{\href{https://github.com/databrickslabs/dolly}{github.com/databrickslabs/dolly}} & \makecell[l]{\citet{liu-reviewergpt-2024}}  \\
			\hline
			\makecell[l]{Gemini 1.5 \\ \citep{gemini-gemini-2024}}
			& \makecell[l]{\usym{2717}} & MoE-Dec & \makecell[l]{\href{https://gemini.google.com/}{gemini.google.com/}} & \makecell[l]{\citet{du-llms-2024},\\ \citet{tyser-ai-2024}}   \\
			\hline
			\makecell[l]{GPT-4 \\ \citet{openai-gpt4-2023}}
			& \makecell[l]{\usym{2717}} & Dec-Only & \makecell[l]{\href{https://openai.com/gpt-4/}{openai.com/gpt-4/}} & \makecell[l]{\citet{bougie-generative-2024},\\ \citet{chamoun-automated-2024},\\   \citet{darcy-marg-2024}, \\ \citet{diaz-streamlining-2024},\\ \citet{du-llms-2024}, \\ \citet{guo-automated-2024},\\ \citet{jin-agentreview-2024},\\ \citet{liu-reviewergpt-2024}, \\ \citet{ong-gpt4-2023},\\ \citet{robertson-gpt4-2023},\\ \citet{shcherbiak-evaluating-2024},\\  \citet{thelwall-can-2024},\\ \citet{tyser-ai-2024}, \\ \citet{yu-automated-2024},\\ \citet{zhou-is-2024}} \\
			\hline
			\makecell[l]{GPT-4o \\ \citep{openai-gpt4-2023}}
			& \makecell[l]{\usym{2717}} & Dec-Only & \makecell[l]{\href{https://openai.com/index/hello-gpt-4o/}{openai.com/index/hello-gpt-4o/}} & \makecell[l]{\citet{bougie-generative-2024},\\ \citet{idahl-openreviewer-2024}}  \\
			\hline
			\makecell[l]{GPT-4o-mini \\ \citep{openai-gpt4-2023}}
			& \makecell[l]{\usym{2717}} & Dec-Only & \makecell[l]{\href{https://openai.com/index/gpt-4o-mini-advancing-cost-efficient-intelligence/}{openai.com/index/gpt-4o-mini-advancing-cost-efficient-intelligence/}} & \makecell[l]{\citet{bougie-generative-2024}}  \\
			\hline
			\makecell[l]{Koala \\ \citep{geng-koala-2023}}
			& \makecell[l]{\usym{2713}} & Dec-Only & \makecell[l]{\href{https://bair.berkeley.edu/blog/2023/04/03/koala/}{bair.berkeley.edu/blog/2023/04/03/koala/}} & \makecell[l]{\citet{liu-reviewergpt-2024}}  \\
			\hline
			\makecell[l]{Llama 2- 70B \\ \citep{touvron-llama2-2023}}
			& \makecell[l]{\usym{2713}} & Dec-Only & \makecell[l]{\href{https://www.llama.com/docs/model-cards-and-prompt-formats/other-models/\#meta-llama-2}{www.llama.com/docs/model-cards-and-prompt-formats/other-models/\#meta-llama-2}} & \makecell[l]{\citet{tyser-ai-2024}}  \\
			\hline
			\makecell[l]{Llama 3-8B  \\ \citep{touvron-llama-2023}}
			& \makecell[l]{\usym{2713}} & Dec-Only & \makecell[l]{\href{https://www.llama.com/docs/model-cards-and-prompt-formats/other-models/\#meta-llama-3}{www.llama.com/docs/model-cards-and-prompt-formats/other-models/\#meta-llama-3}} & \makecell[l]{\citet{du-llms-2024}}  \\
			\hline
			\makecell[l]{Llama 3-70B  \\ \citep{touvron-llama-2023}}
			& \makecell[l]{\usym{2713}} & Dec-Only & \makecell[l]{\href{https://www.llama.com/docs/model-cards-and-prompt-formats/other-models/\#meta-llama-3}{www.llama.com/docs/model-cards-and-prompt-formats/other-models/\#meta-llama-3}} & \makecell[l]{\citet{du-llms-2024}}  \\
			\hline
			\makecell[l]{Llama 3.1-8B  \\ \citep{llama-llama3-2024}}
			& \makecell[l]{\usym{2713}} & Dec-Only &\makecell[l]{\href{https://www.llama.com/docs/model-cards-and-prompt-formats/llama3_1/}{\texttt{www.llama.com/docs/model-cards-and-prompt-formats/llama3\_1/}}} & \makecell[l]{\citet{bougie-generative-2024},\\ \citet{idahl-openreviewer-2024}}  \\
			\hline
			\makecell[l]{Llama 3.1-70B  \\ \citep{llama-llama3-2024}}
			& \makecell[l]{\usym{2713}} & Dec-Only & \makecell[l]{\href{https://www.llama.com/docs/model-cards-and-prompt-formats/llama3_1/}{\texttt{www.llama.com/docs/model-cards-and-prompt-formats/llama3\_1/}}} & \makecell[l]{\citet{bougie-generative-2024},\\ \citet{idahl-openreviewer-2024}}  \\
			\hline
			\makecell[l]{Llama-OpenRev\\iewer-8B \\ \citep{idahl-openreviewer-2024}}
			& \makecell[l]{\usym{2713}} & Dec-Only & \makecell[l]{\href{https://huggingface.co/maxidl/Llama-OpenReviewer-8B}{huggingface.co/maxidl/Llama-OpenReviewer-8B}} & \makecell[l]{\citet{idahl-openreviewer-2024}}  \\
			\hline
			\makecell[l]{Mistral-7B  \\ \citep{jiang-mistral-2023}}
			& \makecell[l]{\usym{2713}} & Dec-Only & \makecell[l]{\href{https://mistral.ai/news/announcing-mistral-7b}{mistral.ai/news/announcing-mistral-7b}} & \makecell[l]{\citet{bougie-generative-2024}}  \\
			\hline
			\makecell[l]{Open Assistant \\ \citep{kopf-openassistant-2023}}
			& \makecell[l]{\usym{2713}} & Dec-Only & \makecell[l]{\href{https://github.com/LAION-AI/Open-Assistant}{github.com/LAION-AI/Open-Assistant}} & \makecell[l]{\citet{liu-reviewergpt-2024}}  \\
			\hline
			\makecell[l]{Qwen2-70B \\ \citep{yang-qwen2-2024}}
			& \makecell[l]{\usym{2713}} & Dec-Only & \makecell[l]{\href{https://qwenlm.github.io/blog/qwen2/}{qwenlm.github.io/blog/qwen2/}} & \makecell[l]{\citet{du-llms-2024}}   \\
			\hline
			\makecell[l]{StableLM \\ \citep{stability-stability-2023}}
			& \makecell[l]{\usym{2713}} & Dec-Only & \makecell[l]{\href{https://github.com/stability-AI/stableLM}{github.com/stability-AI/stableLM}} & \makecell[l]{\citet{liu-reviewergpt-2024}}  \\
			\hline
			\makecell[l]{Vicuna \\ \citep{chiang-vicuna-2023}}
			& \makecell[l]{\usym{2713}} & Dec-Only & \makecell[l]{\href{https://vicuna.lmsys.org}{vicuna.lmsys.org}}&\makecell[l]{\citet{liu-reviewergpt-2024}}  \\
			\hline
		\end{tabular}
	\end{adjustbox}
\end{table}

\begin{figure*}
	\caption{Occurrence distribution of LLMs in ASPR applications.\label{fig:llms-collection-aspr}}
	\includegraphics[width=0.94\textwidth]{fig-used-llms-percentage}
\end{figure*}

The pie chart of ``Model Series'' illustrates the distribution of model series in the ASPR field, divided into three main categories: GPT series, Llama series, and Others. Among them, the GPT series, including ChatGPT, GPT-4, GPT-4o, and GPT-4o-mini, occupies a dominant position with a significant lead. This dominance can be attributed to three factors: outstanding benchmark performance, highly optimized interaction design, and an active developer ecosystem. In contrast, the Llama series, which includes Llama 2, Llama 3, and other iterative versions, holds a relatively smaller proportion, indicating that there is potential for applicability improvement in the ASPR field. Remarkably, the ``Others'' category accounts for over a third of the total, highlighting the diversity of ASPR research tools. This indicates the flexibility available to users in choosing non-mainstream models that are better suited to their particular requirements.

\section{New technologies in the era of LLM-driven ASPR}

\label{sec:new-tech}

In our previous work on ASPR, we reviewed existing technologies that can be leveraged to achieve ASPR, identified existing challenges, and outlined the future potential of utilizing LLMs for this purpose~\citep{lin-automated-2023}. Since then, the advent of LLMs has introduced numerous emerging technologies that have the potential to address the challenges previously faced by ASPR.

\subsection{Long text modeling}

Scholarly paper lengths can vary widely depending on the field, publication type, and purpose. The number of words may fluctuate between several thousand to over a hundred thousand. However, the typical processing capacity of conventional methods and earlier pre-trained language models is often limited to fewer than 6k of tokens (1 token approximately equal to 0.75 word). To address this limitation, methods such as the extract-then-generate method~\citep{yuan-can-2022} and modular guided method~\citep{lin-moprd-2023} have been proposed. Nonetheless, these methods remain insufficient as they are unable to consider the full context and understand across the entire text.

This landscape changed in June 2023, during the era of LLMs as the gpt-3.5-turbo-16k model shows its capability to process 16k input tokens from the beginning. In a short period, the accepted input length has increased to 32k, and then to 64k, a capability that is sufficient for processing most scholarly papers. GPT-3.5-turbo-16k can now process an entire scholarly paper in a single pass. This advancement lays the technical foundation for further development of ASPR.

\subsection{Multimodal input}

Apart from text, scholarly papers can contain a variety of other elements that support the research and help convey complex information. These visual elements include graphs, charts, tables, figures, diagrams, images, photographs, maps, etc. The ability to handle these diverse elements is crucial for ASPR, as it allows models to assess the full breadth of a scholarly paper. The practice of ASPR was primarily limited to textual data processing while overlooking critical non-textual data. This limitation made ASPR less reliable, as it failed to account for the full range of information contained in scholarly papers.

However, with the advent of multimodal LLMs capable of processing both textual and non-textual data~\citep{yin-survey-2024,zhang-mm-2024}, this limitation is being addressed. By incorporating both textual and visual data into the review process, multimodal LLMs enable machines to act fully as human reviewers to review the whole content of a paper, such as text, tables, figures, and supplementary resources including demonstrative audio, demonstrative videos, linked websites, and corresponding source code. The multimodal capabilities open up new possibilities for more efficient, accurate, and scalable evaluations of scholarly papers.

\subsection{Multi-turn conversation}

In the traditional peer review process, reviewing a paper involves a multi-turn pipeline, where authors, reviewers, and editors engage in multi-turn discussions to refine and enhance the quality of the manuscript before publication. In our previous paper, we defined ASPR as a one-turn process~\citep{lin-automated-2023}. Because at that time, multi-turn, specifically long-text multi-turn conversations, in NLP research were not yet sufficiently developed.

As a breakthrough from the earlier models, LLMs have significantly advanced to handle multi-turn long-text conversations. This is achieved primarily through sending the previous inputs into the following conversation, combined with their advanced capabilities of long-text processing. LLMs are now able to engage in multiple turns of conversations to simulate a full scholarly paper review process, including author responses and interactions between reviewers~\citep{tan-peer-2024}. Therefore, LLMs can better support ASPR by evaluating the author responses, checking their revisions, and providing iterative feedback.

\subsection{Instant knowledge acquisition}

In the peer review process, reviewers may encounter descriptions or clarifications beyond their knowledge scope. This can occur if the content is based on the latest updates or research advancements, or if the content falls outside the reviewers' area of expertise. In such cases, the reviewers would search for, verify, and confirm the information with which they are unfamiliar. In the early stage of LLMs, the knowledge base is frozen at the time of training. For example, the knowledge of GPT-4o was updated to October 2023.\footnote{https://platform.openai.com/docs/models/gpt-4o. Data accessed in May 2025.} This is known as the ``knowledge cutoff date''. While providers of LLMs strive for regular model updates, the pace of news and research development far outstrips the update cycle. Even with weekly updates, LLMs cannot fully keep up with real-time information. Moreover, training an LLM requires a significant amount of resources and costs, rendering frequent and timely updates impractical.

To overcome this hurdle, providers of LLMs have integrated search engine function into their models, such as Kimi\footnote{https://kimi.moonshot.cn/} and ChatGPT.\footnote{https://openai.com/index/introducing-chatgpt-search/} When a model's knowledge is outdated or insufficient, it can perform real-time searching on the internet to produce a more up-to-date and accurate answer. This feature allows LLMs to stay aligned with the latest scientific research and to evaluate the ideas and originality of papers more comprehensively.

\section{New methods for generating review reports by LLMs}

\label{sec:new-method}

Review reports are the core output of both peer review and ASPR. In the era of LLM-driven ASPR, there are mainly three types of methods to generate this core output.

\subsection{Prompt engineering}

Prompt engineering refers to the process of designing and optimizing textual prompts for input to LLMs, which are tailored to guide LLMs to produce high quality and expectation-aligned output~\citep{marvin-prompt-2023,sahoo-systematic-2024}. By refining prompts based on specific requirements, researchers can encourage LLMs to generate more accurate, comprehensive, and format-compliant review reports.

The work of \citet{robertson-gpt4-2023} appears to be the first research paper to employ LLMs for large-scale generation of review comments using detailed prompts. These prompts effectively facilitate the generation of review comments across various evaluation aspects while maintaining a consistent format. \citet{liang-can-2024} present an early and inspiring work on creating an automated pipeline for review report generation. The workflow starts by parsing a scholarly paper from PDF to text format, followed by inputting the parsed text along with a paper-specific prompt covering aspects of significance, novelty, acceptance or rejection rationale, and improvement suggestions into LLMs for generation. \citet{thelwall-can-2024} uses the customized function of OpenAI to design a self-defined ChatGPT-4 based on the UK Research Excellence Framework (REF) 2021.\footnote{https://2021.ref.ac.uk/publications-and-reports/index.html} This model is applied to evaluate 51 of his own papers, with the generated scores being compared to his personal quality judgments. \citet{du-llms-2024} utilize a long prompt, consisting of ICLR review guidelines, a selection of randomly chosen accepted and rejected paper reviews written by humans, and an ICLR 2024 review format template, to generate LLM reviews. \citet{liu-reviewergpt-2024} design prompts for LLMs to identify errors in papers, verify papers with checklist questions, and select the ``better'' paper from paired comparisons. \citet{zhou-may-2024} summarize the review criteria of leading cell biology journals as originality, accuracy, conceptual advancement, timeliness, and significance. Then these criteria are used to prompt the LLM to evaluate a given PubMed ID paper on a three-star scoring system for a final overall assessment. \citet{zhou-is-2024} evaluate the ability of LLMs in score prediction and review comment generation. Additionally, they construct a review-revision multiple-choice question dataset (RR-MCQ) to check the LLMs' performance in answering multiple-choice questions related to the reviewed papers.

\subsection{Supervised fine-tuning}

Supervised fine-tuning is the further training process where a pre-trained model is refined through additional training on labeled data to improve its performance for a specific task or domain~\citep{han-parameter-2024,wang-parameter-2024}. Researchers create and compile new datasets to fine-tune LLMs for review report generation.

\citet{tan-peer-2024} introduce ReviewMT, a dataset with comprehensive annotations for each turn during the peer review process. They propose converting the peer review process into a multi-turn dialogue, where different LLMs assume the roles of authors, reviewers, and decision-makers. By fine-tuning LLMs for these roles, they present metrics for assessing text quality, scoring, and decision making, which provides a new perspective for evaluating the performance of LLMs in reviewing. \citet{gao-reviewer2-2024} develop REVIEWER2, a two-stage review generation system utilizing two fine-tuned language models. The first analyzes papers and generates prompts of aspects to be considered in the review, while the second generates specific reviews based on these prompts. In addition, they propose two new evaluation metrics for measuring the specificity and coverage of the generated reviews. \citet{faizullah-limgen-2024} construct LimGen, a dataset comprising 4,068 research papers and their corresponding limitations. They use this dataset to fine-tune LLMs with the non-truncated, dense passage retrieval, and chain modeling methods. Their approach is effective in generating limitations for a given paper by considering its entire context.

\subsection{Multi-agent frameworks}

Prompt engineering and supervised fine-tuning are relatively straightforward methods for review report generation, where a single LLM agent is used. Not satisfied with the performance of single-LLM-agent generation, some researchers combine multiple LLM agents and design special strategies to enable different agents to work together effectively for generation.

\citet{jin-agentreview-2024} introduce AgentReview, a framework that models the conference peer review process using LLM agents. It simulates interactions among authors, reviewers, and area chairs (ACs) across five phases: reviewer assessment, reviewer-author discussion, reviewer-AC discussion, meta-review compilation, and AC decision-making. Review reports are generated and revised throughout these phases. AgentReview is claimed by the authors to be the first work that utilizes LLMs to simulate the peer review process. The study also finds that changes in the review reports are influenced by factors aligned with foundational sociological theories. \citet{tyser-ai-2024} develop AI-driven review systems with two main methods proposed. One is to combine fix questions with adaptive LLM-generated questions to prompt LLMs for reviewing. This method offers greater customization and allows for more paper-specific inquiries. The other is to assign LLMs the roles of program chair, senior area chair, area chair, reviewers, and authors. By simulating a real conference paper submission process, this method seeks to increase the quality of review reports. \citet{chamoun-automated-2024} propose SWIF$^2$T, a system designed to deliver specific, actionable, and coherent comments on scientific papers by leveraging multiple LLMs to identify weaknesses or suggest improvements. The system consists of four components: a planner, an investigator, a reviewer, and a controller. SWIF$^2$T is innovative in that it breaks down the process of generating scientific feedback into a series of steps, and incorporates a plan rescheduling methodology to optimize coherence, structure, and specificity, thereby improving the effectiveness of the scientific feedback generation process. \citet{yu-automated-2024} present SEA, a framework consisting of three modules: Standardization, Evaluation, and Analysis. The Standardization module harnesses GPT-4 to consolidate multiple review comments into a uniformly formatted review. The Evaluation module then fine-tunes the standardized data to generate high-quality, structured reviews. The Analysis module introduces a new assessment metric, the mismatch score, to evaluate the consistency between the content of the paper and the review comments. By integrating these modules, SEA establishes a review process capable of delivering practical and specific comments. \citet{darcy-marg-2024} create MARG, short for multi-agent review generation, a method that employs multiple LLM agents for internal discussions to generate reviews for scientific papers. The agents are categorized into three roles: leader agents, responsible for task coordination and communication; worker agents, assigned to certain sections of a paper; and expert agents, specialized in specific sub-tasks to support the leader agent. With the multi-agent architecture, context management strategies, and refinement stages, MARG can produce more specific, useful, and accurate reviews of scientific papers. Building on a similar multi-agent architecture, \citet{taechoyotin-mamorx-2024} introduce MAMORX, a system designed for multimodal scientific review generation. MAMORX assigns specialized agents to assess various aspects of a paper, such as novelty, figure critique, and clarity. Apart from textual content, the system incorporates multimodal inputs, including figures, tables, and citations. Further, it enhances review quality and accuracy by accessing latest research advances and domain knowledge from Semantic Scholar.

\section{New datasets}

\label{sec:new-dataset}

Datasets are integral to the process of ASPR. As mentioned earlier, they are essential to supervised fine-tuning and serve as the foundation for evaluating task performance. The quality of these datasets directly impacts the accuracy of the review outcomes and the reliability of the review reports. In our previous work, we conducted a comprehensive review of datasets for ASPR~\citep{lin-automated-2023}. Since then, there have been several newly constructed datasets designed to establish benchmarks for tasks relying on traditional ASPR methods. Moreover, with the advent of LLMs, many researchers have constructed new datasets specifically tailored to tasks enabled by these advanced models.
Table~\ref{tab:new-datasets} provides an updated review of recently developed datasets that have emerged since our previous study.

\begin{table}[H]
    \centering
    \caption{New ASPR-related datasets}
    \label{tab:new-datasets}

    \setcellgapes{1.5pt}
    \makegapedcells
    \small
    
    \begin{adjustbox}{width=0.95\textwidth,keepaspectratio}

    \begin{tabular}{@{}l@{\hspace{1pt}}l@{\hspace{1pt}}l@{\hspace{2pt}}l@{\hspace{2pt}}l@{}}
    \hline
    Phase & Name & Application & Scale & Contents  \\ \hline

        Screening
            & \makecell[l]{RelevAI-Reviewer \\ \citep{couto-relevai-2024}} & Relevance assessment & 25k+ instances &  \makecell[l]{Prompts, candidate papers, \\relevance levels, titles, abstracts, \\AI-generated related work sections} \\
        \hline

        \multirow{8}{*}{ \makecell[l]{Main\\review}}
            & \makecell[l]{ArgSciChat\\ \citep{ruggeri-dataset-2023}} & Literature understanding &  \makecell[l]{41 dialogues \\on 20 papers}  &  \makecell[l]{Dialogues, rationales, \\intent classes, abstracts, \\introduction sections}  \\
            & \makecell[l]{ContraSciView\\ \citep{kumar-when-2023}} & Contradiction identification & \makecell[l]{8,582 papers,\\28k+ review pairs,\\50k+ review pair conmments} & \makecell[l]{Papers, review pairs, \\review pair comments} \\
            & \makecell[l]{LimGen\\ \citep{faizullah-limgen-2024}} & Limitation generation & 4,068 papers & Papers, limitation sections \\
            & \makecell[l]{ReviewMT\\ \citep{tan-peer-2024}} & Multi-turn dialogue simulation & \makecell[l]{26k+ papers, \\92k+ reviews} & \makecell[l]{Full text; \\multi-turn reviews} \\
        \hline

        \multirow{3}{*}{ \makecell[l]{Comment\\generation}}
             & \makecell[l]{PEERSUM\\ \citep{li-summarizing-2023}} & Meta-review generation & 14k+ samples  & \makecell[l]{Abstracts, metadata, \\official reviews, public reviews , \\official responses, public responses, \\author responses, author comments} \\
             & \makecell[l]{Reviewer2 Dataset\\ \citep{gao-reviewer2-2024}} & Feedback generation & \makecell[l]{27k+ papers, \\99k+ reviews} & \makecell[l]{Papers, metadata, \\reviews, prompts} \\
        \hline

        \multirow{16}{*}{ \makecell[l]{Review\\quality\\assessment}}
            & \makecell[l]{PolitePEER\\ \citep{bharti-politepeer-2024}} & Politeness of reviews & \makecell[l]{2500 labeled sentences,\\70k unlabeled sentences}  & Review sentences, politeness levels  \\
            & \makecell[l]{SubstanReview\\ \citep{guo-automatic-2023}} & Substantiation of reviews & 550 reviews  & Claims, evidence, SubstanScore \\
            & \makecell[l]{Dataset of \\ \citet{latona-ai-2024}} & \makecell[l]{Impact on submission scores \\and acceptance rates} & \makecell[l]{23k+ submissions,\\86k+ reviews}   & \makecell[l]{Titles, abstracts,  author institutions, \\keywords, introduction sections,\\textual evaluations, \\confidence scores, overall ratings} \\
            & \makecell[l]{ReviewCritique\\ \citep{du-llms-2024}} & \makecell[l]{Quality of LLM-generated \\reviews, effectiveness of \\issue identification} & 100 papers &  \makecell[l]{Candidate papers, \\human-written reviews, \\LLM-generated reviews, \\ labels and explanations} \\
            & \makecell[l]{RR-MCQ\\ \citep{zhou-is-2024}} & \makecell[l]{Ability of score prediction \\and review generation} & 196 questions & Multiple-choice questions, labels \\
            & \makecell[l]{SchNovel\\ \citep{lin-evaluating-2024}} & \makecell[l]{Capability of novelty \\assessment} & 15k paper pairs &  \makecell[l]{Fields, paper pairs, \\publication years, paper IDs,\\ labels, titles,  abstracts,  metadata} \\
        \hline

        \multirow{3}{*}{ \makecell[l]{Author\\response}}
            & \makecell[l]{JITSUPEER\\ \citep{purkayastha-exploring-2023}} & Rebuttal generation & 2,332 sentences & \makecell[l]{Review sentences, \\attitude roots, attitude themes, \\canonical rebuttals, rebuttal actions} \\
            & \makecell[l]{ARIES\\ \citep{darcy-aries-2024}} & Paper revision & 4,096 comments & \makecell[l]{Papers, peer review feedback, \\author responses, revision history} \\
        \hline\noalign{\smallskip}

        \multirow{7}{*}{\makecell[l]{Multi-\\purpose}}
            & \makecell[l]{MOPRD\\ \citep{lin-moprd-2023}} & \makecell[l]{Meta-review generation;\\editorial decision prediction;\\author rebuttal generation;\\scientometric analysis} & 6,578 papers &  \makecell[l]{Paper metadata, \\manuscript versions, \\review comments, meta-reviews, \\rebuttal letters, editorial decisions} \\
            & \makecell[l]{NLPeer\\ \citep{dycke-nlpeer-2023}} & \makecell[l]{Review score prediction;\\pragmatic labeling;\\guided skimming} &  \makecell[l]{5,672 papers, \\11k+ reports}  & \makecell[l]{Peer reviews, paper drafts, \\revision history, metadata} \\
            & \makecell[l]{Dataset of \\ \citet{severin-relationship-2023}} & \makecell[l]{Relationship between peer \\review characteristics and \\impact factor} & 10k reports & \makecell[l]{Review reports, review metadata, \\impact factor, length, categories, \\reviewer demographics } \\
         \hline

    \end{tabular}
    \end{adjustbox}
\end{table}

As shown in Table~\ref{tab:new-datasets}, these datasets can be categorized into different groups based on specific phases of the review process: screening, main review, comment generation, review quality assessment, and author response. Datasets that support multiple phases are classified under a separate multi-purpose category. For screening, main review, and comment generation, the majority of the datasets contain information about the textual content of the manuscript, metadata, decision-making results, review comments, and reports. Some datasets also capture multi-turn conversations that occur throughout the process of reviewing, enabling LLMs to leverage their strengths in handling complex and iterative discussions. Regarding the task of review quality assessment, certain datasets within this category include manually annotated labels provided by domain experts. These labels help evaluate the quality of review comments across different dimensions.

Overall, the existing ASPR-related datasets remain predominantly confined to textual data, overlooking other multimodal elements, such as figures, images, and tables from the manuscripts. This lack of multimodal integration restricts the potential for developing multimodal LLMs for ASPR. Additionally, operational support for multimodal LLM-driven ASPR is still underdeveloped, restricting the practical application of these models in real-world scenarios.

\section{Source code and online systems for LLM-driven ASPR}

\label{sec:new-code-system}

Source code is critical for researchers and application developers, as it enables the reproduction of the methods and results presented in AI-related research papers~\citep{lin-automatic-2022}. This is especially crucial in the context of ASPR. Unlike theoretical research, ASPR research relies heavily on empirical experiments and without access to the underlying source code, reproducing these experiments becomes nearly impossible. Additionally, when developing a comprehensive ASPR pipeline, leveraging existing source code can significantly reduce the effort required for replication. While some studies have yet to release their source code, efforts are conducted to provide online systems that allow users to experience the latest ASPR technologies. For instance, \citet{tyser-ai-2024} have developed three online review systems designed to provide feedback on scholarly papers, perform trend analysis, and improve the quality of reviews. Table~\ref{tab:code-using-llm} presents a summary of LLM-driven ASPR-related studies that have made their source code publicly available.

\begin{table}[htb]
    \centering
    \caption{Source code for LLM-driven ASPR}
    \label{tab:code-using-llm}

    \setcellgapes{2pt}
    \makegapedcells
    \small
    
    \begin{adjustbox}{width=0.95\textwidth,keepaspectratio}

    \begin{tabular}{@{}l@{\hspace{2pt}}l@{\hspace{2pt}}l@{\hspace{2pt}}l@{\hspace{4pt}}l@{}}
    \hline
    Phase & Approach & Application & Official website  \\ \hline

	Screening
            & \makecell[l]{Li et al. \\ \citep{li-evaluating-2024}} & Abstract screening  & https://github.com/mikeli380/LLMAbstractScreening \\
        \hline
        
        \multirow{10}{*}{ \makecell[l]{Main\\review}}
            & \makecell[l]{RAG-Novelty\\ \citep{lin-evaluating-2024}} & Novelty assessment  & https://github.com/ethannlin/SchNovel \\
            & \makecell[l]{RelevAI-Reviewer\\ \citep{couto-relevai-2024}} & Relevance assessment  & https://github.com/paulohenriquecrs/RelevAI-Reviewer \\
            & \makecell[l]{SciLitLLM\\ \citep{li-scilitllm-2024}} & Literature understanding  & https://github.com/dptech-corp/Uni-SMART/tree/main/SciLitLLM \\
            & \makecell[l]{SLG\\ \citep{faizullah-limgen-2024}} & Limitation examination  & https://github.com/arbmf/LimGen \\
            & \makecell[l]{Tan et al. \\ \citep{tan-peer-2024}} & Process enhancement  & https://github.com/chengtan9907/ReviewMT \\
        \hline

        \multirow{16}{*}{ \makecell[l]{Comment\\generation}}
            & \makecell[l]{ChatReviewer \\ \citeyear{ni-chatreviewer-2023}} & Review generating & https://github.com/nishiwen1214/ChatReviewer \\
            & \makecell[l]{RAMMER\\ \citep{li-summarizing-2023}} & Meta-review generation  & https://github.com/oaimili/PeerSum  \\
            & \makecell[l]{ARIES\\ \citep{couto-relevai-2024}} & Automatic revision  & https://github.com/allenai/aries \\
            & \makecell[l]{MAMORX\\ \citep{taechoyotin-mamorx-2024}} & Review generation  & https://github.com/sciosci/mamorx-review-system \\
            & \makecell[l]{MARG\\ \citep{darcy-marg-2024}} & Feedback generation  & https://github.com/allenai/marg-reviewer \\
            & \makecell[l]{Reviewer2\\ \citep{gao-reviewer2-2024}} & Feedback generation  & https://github.com/ZhaolinGao/Reviewer2 \\
            & \makecell[l]{SEA\\ \citep{yu-automated-2024}} & Review generation  & https://github.com/ecnu-sea/sea \\
        \hline

         \makecell[l]{Author\\response}
            & \makecell[l]{ARIES\\ \citep{couto-relevai-2024}} & Automatic revision  & https://github.com/allenai/aries \\
        \hline

    \end{tabular}
    \end{adjustbox}
\end{table}

As shown in Table~\ref{tab:code-using-llm}, researchers have developed a variety of approaches to address subtasks at different stages within the ASPR process, including screening, main review, comment generation and author response. These approaches leverage the strengths of LLMs in processing long-text multi-turn conversation, outperforming traditional NLP baseline models in most cases. Notably, at the end of 2024, the first open-source multi-module LLM-driven ASPR system, MAMORX~\citep{taechoyotin-mamorx-2024}, was introduced. By integrating textual, graphical, and citation concerns with external knowledge sources, MAMORX effectively simulates key aspects of human review. This comprehensive approach gives it a strong advantage over both human reviewers and baseline models.

The reproducibility of the source code and online systems is also a critical consideration. The SEA project exemplifies best practices in this regard. The authors not only provide full access to the model's architecture, model's weights, and source code, but also offer comprehensive project requirements, installation instructions and usage tutorials. Furthermore, the authors provide demos of the project and engage with users regarding hardware support in the GitHub Issues board. These efforts collectively demonstrate a strong commitment to reproducibility, making it an excellent example of replicable research in the context of LLMs for ASPR.

\section{ASPR enhancement with LLMs}

\label{sec:llm-performance}

Early implementations of ASPR relied on traditional automation tools, focusing primarily on format validation, basic language proofreading, and plagiarism detection. At that stage, review reports were typically generated using predefined templates and pre-trained language models. However, with the emergence of LLMs and the associated technological advancements, the development of new generation methods, datasets, and open source code has significantly improved the performance and functionality of LLM-driven ASPR.

\subsection{Efficiency advancement}

As with many other domains, the integration of LLMs into ASPR has led to a remarkable enhancement in both effectiveness and efficiency. One of the most notable impacts of LLMs is their ability to significantly accelerate the peer review process by rapidly evaluating the suitability of submitted manuscripts and identifying potential issues or ethical concerns. This capability contributes to streamlining the entire publication pipeline~\citep{biswas-chatgpt-2023}.

In the coexistence phase of ASPR and peer review, LLMs also play a crucial role in substantially alleviating the workload of human reviewers. For instance, the experiment conducted by \citet{robertson-gpt4-2023} demonstrates that authors rate LLM-generated review comments as equally helpful as those from human reviewers, with an average score of 3 out of 5. This suggests that authors view LLMs as both valuable and effective in facilitating ASPR. The ongoing development and refinement of LLMs are poised to enhance the efficiency and robustness of the reviewing process, thereby saving researchers' precious time and increasing overall satisfaction in the academic publishing ecosystem~\citep{kuznetsov-what-2024}.

\subsection{Summarization improvement}

In a review report, the summary section typically appears at the beginning and serves to provide a concise overview of the manuscript~\citep{lin-automated-2023}. LLMs have demonstrated remarkable summarization capabilities in generating high-quality summaries due to their advanced underlying technology.

Unlike traditional extractive summarization methods, LLMs primarily employ abstractive summarization for generation. This approach allows them to capture the essential points without directly copying from the reviewed manuscript, demonstrating advanced text comprehension and synthesis skills~\citep{du-llms-2024}. This capability has also been examined by \citet{thelwall-can-2024} and has enabled LLMs to generate ``quality evaluation rationales'' that meet the REF criteria. In addition, the study by \citet{pu-summarization-2023} provides empirical evidence that LLM-generated summaries exhibit great performance in key dimensions such as fluency, factuality, and flexibility compared to human-written references. They argue that the introduction of LLMs represents a significant transformative shift in the field of text summarization.

\subsection{Screening empowerment}

Screening represents the initial phase in the traditional peer review process. Implementing efficient and accurate screening mechanisms at this stage can substantially reduce the resource burden on subsequent stages of review. LLMs have emerged as valuable tools in this context, functioning as a first-pass filter through assessing a manuscript's alignment with a journal's aims and scope, as well as providing a preliminary evaluation of its quality.

LLMs can assist editors in several key screening tasks, including assessing manuscript quality, detecting potential plagiarism, and evaluating topical relevance~\citep{mollaki-death-2024}. Their utility in these areas has been acknowledged by Dr. Lisa A. Fortier, Editor-in-Chief of JAVMA and AJVR~\citep{fortier-leveraging-2023}. In addition to traditional editorial processes, LLMs can also be applied to preprint screening, which often lacks formal editorial oversight, to help prevent the dissemination of low-quality or misleading research~\citep{hosseini-fighting-2023}. Beyond editorial screening, LLMs also show promise in supporting systematic literature review. The research by \citet{li-evaluating-2024} reports that LLMs achieved up to 90\% overall accuracy in abstract screening tasks, underscoring their potential in literature identification.

\subsection{Checklist validation}

Checklists serve as essential tools to ensure that authors have systematically addressed all the required elements before submitting their scholarly manuscripts to journals or conferences. These checklists typically encompass a wide range of criteria, including content completeness, structural coherence, ethical compliance, and adherence to specific formatting guidelines. By guiding authors through these requirements, checklists help improve the overall quality and consistency of submissions. LLMs can be employed to assess whether a manuscript conforms to the prescribed criteria set by a target venue, thereby assisting authors in identifying omissions or inconsistencies prior to submission.

Recent advancements in LLMs have opened up promising opportunities for automating checklist verification. The study of \citet{liu-reviewergpt-2024} has shown that LLMs achieved an 86.6\% accuracy rate in performing checklist verification, an outcome that is comparable to human-level performance. Notably, half of the reported errors in the study were not due to actual misjudgments by the models but rather because certain functionalities listed on the checklist were not yet supported by the LLMs at the time of the experiment. For example, certain checklist items such as figure information verification were beyond the capabilities of unimodal LLMs at that time. As LLMs continue to evolve, their effectiveness in checklist-based verification is expected to improve substantially.

\subsection{Error identification}

Identifying potential errors is an essential aspect of the review process. LLMs have demonstrated the ability to evaluate the accuracy of scholarly papers and identify a wide range of issues, including both mathematical and conceptual errors, thereby contributing to the overall improvement of manuscript quality.

In an experiment conducted by \citet{liu-reviewergpt-2024}, each of 13 computer science papers was deliberately modified to include one technical error, and LLMs successfully identified the errors in 7 of them. These errors covered topics of bias and fairness, non-parametric regression, sorting, noisy pairwise comparisons, classification, game theory, error correcting codes, optimization, clustering, and distinguishing styles. The results suggest that LLMs can achieve error detection performance comparable to that of human experts. Similarly, \citet{tyser-ai-2024} introduced common writing and reasoning flaws into manuscripts, and compared the review scores assigned by LLMs before and after these modifications. Their findings indicate that LLMs are sensitive to issues of overclaiming, insufficient discussion of limitations, citation problems, and theoretical errors.

\subsection{Comment refinement}

Scholarly papers are the culmination of intensive intellectual and scientific efforts. Given the complexity and depth of such work, it is often challenging for human reviewers to thoroughly review a manuscript and compose accurate review comments within a limited time frame. LLMs can assist in this process by helping reviewers in composing more complete and precise feedback.

A common application involves employing LLMs to refine the reviewer's initial draft. By inputting the original comments into an LLM, reviewers can obtain grammatically correct and well-formatted versions of their original reviews~\citep{nath-ai-2024}. A more advanced application is to leverage LLMs' exposure to a large number of published papers and review reports. This enables LLMs to generate suggestions based on both the manuscript's content and the reviewer's initial comments, thereby supporting the production of more insightful, comprehensive, and analytically robust reviews~\citep{hosseini-fighting-2023}.

\section{Main existing issues}

\label{sec:existing-issue}

In the peer review process, reviewers are expected to provide objective and constructive comments. Although LLMs have made significant advancements in enhancing ASPR, they still face several issues in fully replacing human reviewers at the current stage. In this section, we review the limitations of using LLMs for ASPR and analyze the underlying mechanisms.

\subsection{Insufficient knowledge and comprehension}

Reviewing a manuscript is a complex process that requires in-depth knowledge and profound experience in the relevant field. Justification in research is a truth-conducive process, where the verification of results and interpretation of data are critical components. However, LLMs have not yet reached human-level performance in these areas~\citep{takagi-speculative-2023}. LLMs face technical limitations in understanding highly specialized terminology or the latest research advances. They lack true comprehension of domain-specific expertise and the capacity to contextualize feedback within the nuanced framework of a particular field of study~\citep{biswas-ai-2024}. As a result, LLMs fail to match the ability of human reviewers to evaluate a manuscript's scientific contribution within the broader context of its field and determine whether that contribution warrants acceptance~\citep{saad-exploring-2024}. LLMs are more likely to provide brief and generic comments, indicating that their evaluation is currently limited to an imitative assessment of the text's characteristics, and that they are not yet capable of critically evaluating core elements such as originality and soundness~\citep{zhou-may-2024}. Moreover, LLMs tend to provide out-of-scope comments because their reviews are often generic rather than tailored to the specific content of a manuscript. This superficial approach is aimed at minimizing technical errors~\citep{du-llms-2024}.

The current output mechanisms of LLMs rely on predicting missing words using deep neural networks with billions or even trillions of parameters. These networks are trained through unsupervised machine learning methods on enormous amounts of text from various sources, predominantly open-domain web resources, with academic content constituting only a small fraction~\citep{brown-language-2020}. This limitation significantly hinders the ability of LLMs to comprehensively evaluate sophisticated, cutting-edge topics and niche subjects in depth. In contrast, human reviewers are endowed with cognitive abilities of intuition, perception, empathy, deduction, self-awareness, and social awareness. These capabilities allow them to discern conflicting theories and findings, identify inherent and problematic assumptions, and recognize enduring but not outdated contributions.
Unfortunately, LLMs lack these essential human faculties~\citep{weber-other-2024}, and as a result, LLMs struggle to achieve a profound understanding necessary for high quality ASPR.

\subsection{Biased review}

It is the pursuit of peer review to provide an objective and fair evaluation of a manuscript based exclusively on its quality. However, due to the inherent generation mechanisms, the output of LLMs is subject to many factors, including user prompts and training data~\citep{tan-peer-2024}. These factors can introduce variability in the results, leading to inconsistencies in the quality and reliability of the generated contents. LLMs have a tendency to assign higher scores to most papers~\citep{zhou-may-2024,du-llms-2024}. Research by \citet{thelwall-can-2024} found that LLMs never awarded the lowest score to any reviewed paper, and often middle-range scores were assigned instead. Additionally, OpenAI has acknowledged that its LLM may generate biased content~\citep{openai-gpt4-2023}. These biases, which can include prejudices related to factors such as social class, race, and geography, are inherent in LLMs and may be reproduced or even amplified during their use~\citep{hosseini-fighting-2023}.

These biases are inherited from the training data used to develop LLMs~\citep{leung-best-2023}. Whether acknowledged or not, most of the data LLMs are trained on come from individuals that may unknowingly carry their own biases. These biases include a preference for positive outcomes, varying levels of tolerance or intolerance toward authors with specific demographic profiles, or from certain institutions or countries~\citep{hosseini-fighting-2023}. Furthermore, minority groups or perspectives may be underrepresented or absent from the training data in the LLMs, which can lead to bias when handling content related to these groups or viewpoints. This issue is similar to how low-quality manuscripts and their associated review comments may contribute to biased evaluations when LLMs encounter unfamiliar or rarely seen content.

\subsection{Inaccurate and erroneous comments}

Identifying potential problems accurately in manuscripts and providing constructively insightful comments to the authors are fundamental expectations for all reviewers. At a minimum, reviewers are expected to avoid making incorrect or misleading comments. Numerous studies have evaluated the performance of LLMs in ASPR and have highlighted their tendency to produce inaccurate or even erroneous feedback. One significant issue is that LLMs may ``hallucinate''\citep{xu-hallucination-2024,huang-survey-2025}, fabricating information and presenting it alongside fabricated or unrelated supporting references, thereby creating comments that may appear convincing on the surface but are factually incorrect~\citep{spitale-ai-2023,tyser-ai-2024}. Moreover, they may misinterpret weak or low-quality manuscripts, erroneously portraying them as well-researched and credible~\citep{nath-ai-2024}. In many cases, LLMs tend to accept authors' claims without thoroughly verifying them and may approach the task of identifying a manuscript's strengths as a simple and superficial task of text summarization of the manuscript itself rather than a critical evaluation~\citep{du-llms-2024}.

Beyond these issues, LLMs exhibit limitations in detecting deeper theoretical flaws, identifying missing metrics, and recognizing instances of overstated conclusions~\citep{drori-human-2024}. Their evaluations often lack the critical depth necessary for assessing complex academic arguments. Additionally, when tasked with reviewing the same manuscript multiple times, LLMs frequently generate unstable and inconsistent evaluations. The instability and inconsistency are reflected in fluctuating scores, and varied expressions generated across different rounds of evaluation for the same paper~\citep{thelwall-can-2024}. Further research also reveals that LLMs struggle with assessing writing quality~\citep{du-llms-2024}, and are largely incompetent in accurately scoring the overall correctness of a manuscript's content~\citep{zhou-is-2024}.

\subsection{Compromised data security}

In the traditional peer review process, strict confidentiality serves as a foundational principle. Editors and reviewers are entrusted with the responsibility of safeguarding authors' unpublished manuscripts, ensuring that these materials are not disclosed or misused. This is particularly critical given the serious concerns over plagiarism and intellectual misappropriation that have arisen within the scientific community. Although breaches occasionally occur~\citep{laine-scientific-2017}, the overarching commitment to confidentiality has become a broadly recognized and widely upheld norm across academic disciplines. The integration of LLMs into ASPR introduces new and complex confidentiality challenges that have become a primary area of concern~\citep{conroy-how-2023}.

A key issue arises from the potential that LLMs may unintentionally store, process, or even incorporate confidential manuscript content into their training data. Since many LLMs are continuously fine-tuned or updated using user-provided inputs, there exists a risk that unpublished research materials submitted for scholarly review could become part of future training corpora, thereby compromising confidentiality~\citep{hosseini-fighting-2023}. These confidentiality risks are largely attributable to the current deployment models and operational architectures of LLMs. Training, fine-tuning, and deployment of LLMs require substantial resources, including significant domain expertise and computational power. Therefore, to reduce costs, the general method for common users to access LLMs is through online providers, rather than deploying themselves. In such configurations, users interact with LLMs via online platforms, which necessitate uploading the manuscripts and all other accompanying materials to remote servers managed by third-party providers. During these interactions, all transmitted data may be used for various purposes, including service improvement and further model training.\footnote{https://openai.com/consumer-privacy/} There also exists a possibility that the manuscript content could be disclosed in response to other users' prompts. Such disclosures would constitute a serious breach of confidentiality and undermine the core ethical principles underpinning the scholarly review system.

\subsection{Limited customization capability}

The ASPR system is explicitly designed to evaluate manuscripts ``for a specific publication venue'', and to align with its ``aims and scopes, review standards, and publication requirements''~\citep{lin-automated-2023}. In theory, this alignment should allow ASPR to adapt its review process to the distinctive expectations of different journals. However, in practice, LLMs, which serve as the core engines for ASPR, exhibit significant limitations in capturing the nuanced and multifaceted editorial visions that define these publication venues~\citep{saad-exploring-2024}. They lack the capacity to internalize the complex and tacit knowledge that editorial boards employ when forecasting academic trends and shaping the intellectual identity of their journals.

Additionally, existing models are equally constrained in their ability to reflect highly individualized review styles and preferences exhibited by human reviewers. While journal-level review criteria are institutionally codified, the actual evaluation of a manuscript is often further shaped by the reviewer's personal expertise, interpretive lens, rhetorical style, and evaluative priorities. These micro-level variations introduce additional layers of complexity into the review process. Present-day LLMs are not yet capable of accurately modeling these personalized dynamics or simulating the implicit reasoning process that underlies individual reviewers' judgments. As a result, a growing phenomenon has emerged where different reviewers utilize LLMs to generate review comments that are strikingly homogeneous, lacking in differentiation, and ultimately characterized by a ``one-size-fits-all'' approach. This uniformity undermines the richness, diversity, and intellectual rigor that peer review is intended to preserve.

\section{Publisher policies on AI-generated content tools in peer review}

\label{sec:publisher-policy}

ASPR has now reached a stage of coexistence with traditional peer review, as many reviewers have begun using LLMs to generate review reports for their own works~\citep{liang-monitoring-2024}. Recognizing this growing trend, major scholarly publishers have established corresponding policies to regulate the use of AI-generated content (AIGC) tools in the peer review process. A summary of these policies is presented in Table~\ref{tab:ai-reviewing-policy}.

\begin{table}[htb]
    \centering
    \caption{Publisher policies on AIGC tools for scholarly review}
    \label{tab:ai-reviewing-policy}

    \setcellgapes{4pt}
    \makegapedcells
    \small

    \begin{tabular}{@{}l@{\hspace{6pt}}l@{\hspace{6pt}}l@{\hspace{6pt}}l@{}}
    \hline
    \multicolumn{2}{l}{Category}  &  Publisher  &  Usage  \\ \hline

       \multicolumn{2}{l}{ \multirow{14}{*}{ \makecell[l]{Prohibited}} }
            & AAAS~\citeyear{aaas-peer-2024} & —— \\
             \multicolumn{2}{l}{} & AIP~\citeyear{aip-ethics-2024} & —— \\
             \multicolumn{2}{l}{} & Cambridge University Press~\citeyear{cambridge-core-2024} & —— \\
             \multicolumn{2}{l}{} & Elsevier~\citeyear{elsevier-use-2024} & —— \\
             \multicolumn{2}{l}{} & Emerald~\citeyear{emerald-review-2024} & —— \\
             \multicolumn{2}{l}{} & IEEE~\citeyear{ieee-become-2024} & —— \\
             \multicolumn{2}{l}{} & MDPI~\citeyear{mdpi-guidelines-2024} & —— \\
             \multicolumn{2}{l}{} & MIT Press~\citeyear{mit-policy-2024} & —— \\
             \multicolumn{2}{l}{} & PLOS~\citeyear{plos-ethical-2025} & —— \\
             \multicolumn{2}{l}{} & Sage~\citeyear{sage-using-2024} & —— \\
             \multicolumn{2}{l}{} & Springer Nature~\citeyear{springer-editorial-2024} & —— \\
             \multicolumn{2}{l}{} & Taylor \& Francis~\citeyear{taylor-guidelines-2024} & —— \\
        \hline

        \multirow{15}{*}{ \makecell[l]{Partially \\ allowed}}
            &\multirow{6}{*}{ \makecell[l]{External \\ AIGC \\ tools}}
                & ACL~\citeyear{acl-faq-2023} & \makecell[l]{Review paraphrase, proof checking,\\ concept explanation} \\
                & \multirow{6}{*}{} & ACM~\citeyear{acm-acm-2023,acm-acm-2024} & \makecell[l]{Quality and readability enhancement\\ (disclosure not required)}  \\
                & \multirow{6}{*}{} & JAMA Network~\citeyear{flanagin-guidance-2023} &  \makecell[l]{All AIGC tools that comply with confidentiality policies \\(name and usage must be declared) }\\
            \cline{2-4}

            &\multirow{4}{*}{ \makecell[l]{In-house \\ AI tools}}
                & Frontiers~\citeyear{frontiers-artificial-2020,frontiers-editorial-2024} &  \makecell[l]{AI tool: AIRA\\Language quality assessment, figure integrity checks, \\plagiarism detection, potential conflict-of-interest identification} \\
                & \multirow{4}{*}{} & Wiley~\citeyear{wiley-role-2025} & \makecell[l]{AI tool: Research Exchange\\Reviewer matching, integrity screening } \\
            \hline
            
       \multicolumn{2}{l}{ \multirow{6}{*}{ \makecell[l]{Unknown}} }
            & Brill & ——\\
            \multicolumn{2}{l}{} & De Gruyter & ——\\
            \multicolumn{2}{l}{} & Harvard University Press & ——\\
            \multicolumn{2}{l}{} & Oxford University Press & ——\\
            \multicolumn{2}{l}{} & World Scientific & ——\\
        \hline

    \end{tabular}
\end{table}

As can be seen in Table~\ref{tab:ai-reviewing-policy}, most of the publishers prohibit reviewers from using AIGC tools to generate or assist in writing review reports. The main reasons for this restriction include:

\begin{itemize}
\item The usage of AIGC tools requires uploading the content of manuscript to external servers, posing a risk of unauthorized data exposure. Such actions could compromise the confidentiality of the review process.

\item Peer review relies on the knowledge and experience of the reviewer. Reviewers are fully accountable for the content of their review reports, and incorporating AI-generated opinions -- whether in part or in full -- undermines this personal responsibility. Additionally, AI-generated assessments may introduce biases, which could negatively impact the quality and fairness of the review process.
\end{itemize}

Some publishers allow editors and reviewers to incorporate AIGC tools to support the review process, provided that they adhere to confidentiality guidelines. ACM authorizes reviewers to apply generative AI or other third-party tools to improve the quality and readability of their review reports without requiring disclosure. ACL enables reviewers to leverage AIGC tools to help understand concepts and polish review comments. JAMA Network approves the use of AIGC tools as a review resource but mandates disclosure, requiring reviewers to specify the tool and its application. Despite variations in AIGC adoption, these publishers share a common restriction that editors and reviewers must not upload manuscripts to platforms that retain data or fail to comply with confidentiality guidelines.

A few publishers have taken a more proactive approach by developing in-house AI tools to enhance the efficiency of the review process. Frontiers introduces the Artificial Intelligence Review Assistant (AIRA), the first advanced AI tool for peer review. Currently, AIRA can generate up to 20 recommendations in seconds, including language quality assessment, figure integrity checks, plagiarism detection, and potential conflict-of-interest identification. Wiley creates the Research Exchange publishing platform to facilitate the sharing of research findings and their real-world impact. This platform also supports peer review process by facilitating reviewer matching and research integrity screening. Its AI capabilities include detecting AI-generated content, verifying researcher identities, identifying forged or manipulated references, and indicating suspicious or problematic language. Although these two publishers integrate AI into their editorial workflows, they do not allow reviewers to use other external AIGC tools for conducting evaluations or making editorial decisions.

There are also publishers that have not publicly stated their policies regarding the use of AIGC tools in the peer review process. Examples of such publishers are classified as ``Unknown'' in Table~\ref{tab:ai-reviewing-policy}, based on the information available to us at the time of writing. While these publishers all state that they adhere to the guidelines of the Committee on Publication Ethics (COPE), the COPE guidelines currently only address the use of AI tools by authors, without providing corresponding provisions for reviewers. As a result, it remains unclear whether reviewers for these publishers are permitted to use AIGC tools when evaluating manuscripts.

\section{Suggestions from academia}

\label{sec:academia-suggestion}

In contrast to the cautious stance of the publishing community, the academic community has been relatively more receptive to the use of LLMs for ASPR. While they share concerns about data security, the academic community in general has maintained an open-minded approach and offered constructive suggestions for the responsible use and further development of LLMs.

\subsection{Ensuring data security}

Academia places equal importance on data security as the publishing industry. It is recommended that materials containing sensitive information or protected data should not be input into LLMs unless appropriate measures have been taken to ensure data security~\citep{hosseini-fighting-2023,mehta-application-2024}. One possible and practical solution to address the data security concerns is to deploy privately hosted LLMs~\citep{conroy-how-2023}. By utilizing privately hosted LLMs, all data, including the confidential ones, can remain within the user's own server, thereby eliminating the need to submit it to external LLM service providers to ensure greater data security. However, this solution presents several challenges. It demands the involvement of technical experts in LLMs and requires substantial hardware resources. Additionally, open-source LLMs often exhibit inferior performance compared to close-source ones.

\subsection{Establishing training system}

Training within a standardized system is generally considered a necessary measure when addressing novel challenges. \citet{mehta-application-2024} suggest that before integrating LLMs into the peer review process, researchers ought to undergo ethics training and familiarize themselves with the possible limitations and biases of the emerging technology. Similarly, \citet{hosseini-fighting-2023} highlight the importance of integrating LLM-related content into peer review training, underscoring both the potential benefits and drawbacks of employing these models for review. This kind of training fosters responsible usage of LLMs and cultivates awareness of their possible weaknesses and biases of LLMs. Given the rapid pace of technological advancements nowadays, they also recommend that the training should be updated regularly to keep pace with the ongoing evolution of LLMs.

\subsection{Promoting usage declaration and transparency}

Unlike publishers who generally prohibit the use of LLMs in the actual peer review process, academia exhibits a more tolerant stance, accepting their use in reviewing under certain conditions. \citet{drori-human-2024}
advocate for explicit self-declaration or machine-generated watermarks when reviewers employ LLMs in their reviewing, aiming to enhance transparency during the process. In a comparable manner, \citet{hosseini-fighting-2023} suggest that besides disclosing the use of LLMs, reviewers should also take full responsibility for the originality, accuracy, and tone of their review reports. Editors are likewise encouraged to acknowledge the involvement of LLMs in the manuscript review process. \citet{mehta-application-2024} further recommend that reviewers and editors share their experiences and outcomes when using LLMs to foster collective learning. \citet{kuznetsov-what-2024} claim that to improve the transparency in LLMs-assisted review, it is essential to clearly outline the tools employed, their functionalities, limitations, and the decision-making processes.

\subsection{Exploring optimal strategies for LLM utilization}

Researchers have accumulated diverse experiences and developed various techniques through their work with LLMs. LLMs are evolving rapidly, often with providers offering multiple versions simultaneously, each varying in performance and cost. Using the latest and most up-to-date version can significantly enhance the integration of LLMs and ASPR~\citep{saad-exploring-2024}. Moreover, given that ASPR involves tasks with varying levels of complexity, adopting simpler models for less demanding tasks can enhance efficiency. Exploring strategies to switch dynamically between models based on task complexity and cost-effectiveness presents a promising research avenue~\citep{darcy-marg-2024}. To achieve more accurate performance, it is recommended to prompt LLMs for outputs over multiple rounds and average the evaluation scores, as this yields more reliable results than relying solely on a single round~\citep{thelwall-can-2024}. To reduce errors, manuscripts should be divided into smaller, more manageable sections before being submitted to LLMs~\citep{mehta-application-2024}. Researchers are also encouraged to continuously experiment with LLMs and share their experiences and findings to overcome the uncertainties with regard to their competencies, limitations, and internal mechanisms~\citep{hosseini-fighting-2023}. Last but not least, researchers should avoid relying exclusively on LLMs for scholarly review~\citep{mehta-application-2024} to prevent potential technical breakdown~\citep{drori-human-2024}.

\subsection{Aligning with academic research values and goals}

Although human reviewers might have different opinions and comments on the same manuscript, academia shares common values and objectives to promote the development of the human world for a better life in the future. \citet{robertson-gpt4-2023} underscores the need for further research to ensure that the LLMs used in the peer review process are consistent with our values and goals for scholarly research. In a similar vein, \citet{takagi-speculative-2023} recommends that when employing LLMs for reviewing manuscripts, it is essential to ensure that their value judgments align with the assessments of human researchers. This alignment helps maintain review comments consistent with human perspectives and values, rather than reflecting the viewpoints or biases inherent to LLMs. Furthermore, \citet{drori-human-2024} propose that LLMs should be monitored and reported on for compliance with journal guidelines, including adherence to review forms and the rules established by editorial and professional organizations.

\section{Challenges and future directions}

\label{sec:concl}

In this survey paper, we review and synthesize existing literature to provide an overview of the technologies, methods, resources, policies, responses, and suggestions related to ASPR within the scope of LLMs. When LLMs first came into existence, they were met with skepticism and resistance from prestigious institutions and universities, as concerns were raised regarding their utility and ethical implications. Over time, however, these models have demonstrated their potential to revolutionize various tasks, ultimately gaining acceptance and becoming integral to modern education systems~\citep{mcdonald-generative-2025,jin-generative-2025}. A similar trajectory is now being observed with ASPR. The idea of automating this traditionally human-led process has initially encountered resistance as it enters the coexistence phase with peer review. Nonetheless, advancements in LLMs have begun to show that ASPR can be achieved in full sense, and we are confident that in the near future, the publishers will adopt ASPR with an open mind, recognizing its value as a standard tool to enhance the efficiency and overall quality of scholarly evaluation.

Based on the insights derived from this review, we highlight several open challenges as follows to warrant further exploration over this topic.

\textbf{Correcting hallucinations.}
The issue of hallucinations~\citep{xu-hallucination-2024,huang-survey-2025} in LLMs has been a persistent challenge since their inception, stemming from the inherent complexities of the generation mechanisms. LLMs generate text based on probabilistic chain decoding, which prioritizes statistical plausibility over factual correctness. As a result, existing evaluation metrics often overemphasize surface-level fluency while neglecting the crucial aspect of factual consistency. Moreover, factors, including noise contamination in the training data, inadequate coverage of knowledge domains, and the fragility of contextual reasoning, further compound the risk of generating fabricated and misleading content. In the context of scholarly paper review, it is crucial to ensure that the generated content remains authentic and reliable for accuracy, clarity, and factual integrity. Despite the ongoing research efforts to devise more robust text generation methods aiming at mitigating hallucinations, these erroneous outputs continue to emerge, particularly in complex scenarios. This persistent challenge underscores the need for further advancements in LLMs to enhance the reliability of generated content. Promising directions include domain adaptation through dynamic knowledge fusion, multimodal fact-checking, and interpretability enhancement. Collectively, these strategies aim to align LLM outputs with the rigorous standards of academic and other critical applications.

\textbf{Generating with multimodal inputs.}
Although LLMs have entered the multimodal era, research on ASPR leveraging multimodal inputs remains relatively limited. A fundamental obstacle lies in the scarcity of high-quality, richly annotated multimodal datasets that offer precise labeling and balanced distributions. Challenges persist across several stages of the dataset development pipeline, including data acquisition, data annotation, data cleaning, and dataset balancing. Without appropriate data, it is difficult for models to fully learn the complex associations and interaction patterns between different modalities, thus constraining advancements of related research. Therefore, it is imperative to develop well-curated multimodal datasets and further conduct ASPR research on them. This represents a highly promising and valuable direction for improvement.

\textbf{Applying reasoning models.}
Reasoning models are a class of LLMs designed to perform systematic reasoning through the explicit construction of chain-of-thought. These models operate by pre-generating a complete reasoning process, decomposing the solution process of a complex problem into intermediate logic nodes, explicitly displaying interpretable steps of reasoning, and ultimately outputting a conclusion verified through multiple steps of inference. This chain-based reasoning architecture substantially enhances the model's problem-solving capabilities in complex scenarios, and has demonstrated notable advantages in tasks such as program debugging, scientific hypothesis testing, and collaborative planning among multiple intelligent agents. Over the past two years, several reasoning models, including OpenAI's o1 series, DeepSeek's DeepSeek-R1, and Google's Gemini, have emerged and found broad application across various fields. However, their application in the field of ASPR remains exploratory. The integration of reasoning models offers the potential to shift ASPR from statistical pattern matching toward an intelligent review paradigm that approximates human-like critical thinking. Their ability to deeply deconstruct argumentative logic can reduce the dependency on surface-level textual features, while the complete thought chains can provide a structured foundation for decision auditing. Nevertheless, this transformation is not without challenges. Potential risks, such as hallucinations induced by excessive reasoning and tensions between timeliness and computational resources, must be carefully managed to ensure reliable application.

\textbf{Mitigating generative attacks.}
Generative attacks pose a significant security challenge for LLMs. In the context of ASPR, their impacts primarily manifest in two areas. First, adversarial attacks occur when malicious actors craft adversarial inputs that induce models to generate review conclusions deviating from the principles of academic integrity, directly undermining the credibility of the peer review process~\citep{ye-are-2024}. For example, an attacker might prompt a model to fabricate references or endorse incorrect claims. Second, sensitive information extraction exploits vulnerabilities in model interpretability, enabling attackers to reconstruct unpublished manuscript content or infer author identities through targeted queries, thereby significantly elevating the risk of data breaches. To ensure a secure and reliable deployment of LLMs in ASPR, a multi-layered defense framework is to be established. At the model level, robustness can be reinforced by detecting atypical inputs and suppressing abnormal behaviors. At the system level, stringent access controls and data flow encryption mechanisms serve as vital protective barriers. At the application level, end-to-end audit trails and proactive early-warning systems can be enforced. By systematically integrating defenses across these three layers, robust safeguards and clear trust boundaries can be established to effectively mitigate the risks associated with generative attacks in ASPR.

\textbf{Deploying low-resource privately hosted models.}
Confidentiality is a primary concern for publishers and researchers utilizing LLMs for ASPR. While privately hosted models offer a viable solution to address data privacy and security, their deployment is often hindered by substantial GPU requirements and sustained computational energy consumption, which significantly increase implementation costs. Current technical efforts to enable low-resource deployment focus on three core dimensions. First, employment strategy optimization aims to reduce per-inference resource demands through techniques such as model compression (e.g.,~pruning, distillation, quantization), hardware acceleration, and the use of caching mechanisms. Second, algorithm enhancement seeks to improve computational efficiency through dynamic resource scheduling algorithms and mixed-precision training. Third, cost-control frameworks incorporate real-time resource monitoring and elastic resource allocation to facilitate distributed deployment. Recent innovations from DeepSeek~\citep{deepseek-v3-2024,deepseek-r1-2025} exemplify and advance these directions. Together, these technological advancements pave the way for practical deployment of privatized ASPR systems by small and medium-sized publishers and even individual reviewers in local environments. They lay the groundwork for secure and controllable intelligent peer review ecosystems, thereby supporting broader adoption of ASPR technologies across the academic landscape.

\textbf{Enabling personalized review.}
The current ASPR system is built upon a standardized framework, which limits its ability to accommodate the individual needs of different disciplines, journals, or reviewers, resulting in a tension between the standardization of review comments and the flexibility required for diverse academic contexts. Future research can address this challenge in the following two directions. One direction is dynamic review standard adaptation, which involves developing mechanisms for dynamic adjustment based on disciplinary characteristics or journal-specific review guidelines. This will enable LLMs to automatically align with the review standards of distinct fields and flexibly adapt to diversified evaluation needs. The other direction is reviewer style modeling and migration, which focuses on constructing personalized models by leveraging historical review data to capture individual reviewers' preferences. Incorporating reinforcement learning feedback mechanisms, where editors or authors rate the generated reviews, can further iteratively optimize the model's alignment with the target review styles. By advancing research in these areas, the ASPR system can maintain the efficiency of automation while preserving the diversity and humanistic values inherent in academic evaluation.

\section*{Declaration of competing interest}
The authors declare that they have no known competing financial interests or personal relationships that could have appeared to influence the work reported in this paper.

\section*{Acknowledgments}
This work is funded by the Research Initiation Fund Project of Guangzhou Institute of Science and Technology (2023KYQ184), the Guizhou Province Science and Technology Plan Project (Qian Ke He [2023] No. 450), the Guangdong Province Key Construction Discipline Research Capacity Enhancement Project (2024ZDJS101), and the Special Innovative Projects of Ordinary Colleges and Universities in Guangdong Province (2024KTSCX139). Special and heartfelt gratitude goes to the corresponding author's wife Fenmei Zhou, for her understanding and love. Her unwavering support and continuous encouragement enable this research to be possible.

\section*{Data availability}
No data was used for the research described in the article.

\bibliography{mybib}

\end{CJK}
\end{document}